\documentclass[11pt]{article}

\usepackage[preprint]{acl}

\usepackage{times}
\usepackage{latexsym}

\usepackage[T1]{fontenc}
\usepackage{listings}   
\usepackage[none]{hyphenat}  
\usepackage{xcolor}     
\usepackage[utf8]{inputenc}
\usepackage[T1]{fontenc}
\usepackage{graphicx}
\usepackage{subcaption}
\usepackage{listings}
\usepackage{listingsutf8}
\usepackage{booktabs}
\usepackage{amsmath,amssymb}

\usepackage{listings}
\usepackage{xcolor}
\usepackage{caption}
\usepackage{todonotes}

\lstdefinestyle{prompt}{
  basicstyle=\ttfamily\footnotesize,
  columns=fullflexible,
  keepspaces=true,
  showstringspaces=false,
  breaklines=true,
  breakatwhitespace=false,
  frame=single,
  framerule=0.2pt,
  framesep=3pt,
  xleftmargin=0pt,
  xrightmargin=0pt,
  resetmargins=true,
  linewidth=\columnwidth
}
\usepackage{listings}
\usepackage{xcolor}
\usepackage{caption}   
\usepackage{varwidth}  
\lstdefinestyle{prompt}{
    basicstyle=\ttfamily\footnotesize, 
    breaklines=true,                   
    breakatwhitespace=true,            
    postbreak=\mbox{\textcolor{gray}{$\hookrightarrow$}\space}, 
    frame=single,                      
    columns=flexible,                  
    xleftmargin=2pt,                   
    xrightmargin=2pt,
    keepspaces=true                    
}

\usepackage{microtype}

\usepackage{graphicx}

\title{FairJudge: Abstention-Aware Multimodal Judges for Fairness and Alignment Evaluation in Text-to-Image Models}

\author{
  Zahraa Al Sahili\textsuperscript{1}\thanks{\ \ Corresponding author.} \quad
  Maimuna Nowaz\textsuperscript{1} \quad
  Maryam Fetanat\textsuperscript{3} \quad
  Ioannis Patras\textsuperscript{1} \quad
  Matthew Purver\textsuperscript{1,2} \\
  \textsuperscript{1}Queen Mary University of London, UK \\
  \textsuperscript{2}Institut Jožef Stefan, Slovenia \\
  \textsuperscript{3}Imperial College London, UK \\
  \texttt{\{z.alsahili, i.patras, m.purver\}@qmul.ac.uk} \\
  \texttt{m.nowaz@se23.qmul.ac.uk} \quad
  \texttt{maryam.fetanat20@imperial.ac.uk}
}

\begin{document}
\maketitle

\begin{abstract}
Evaluating text-to-image (T2I) systems requires judging not only whether an image matches a prompt, but also whether socially salient attributes are represented faithfully and without unsupported inference. Existing automated evaluators typically rely on face-centric recognizers or contrastive image--text similarity, which provide limited diagnostic feedback and often force predictions even when visual evidence is ambiguous or absent. For fairness-sensitive attributes such as religion and disability, where cues may be contextual, indirect, or intentionally unspecified, these evaluators can therefore miss failure modes that careful human reviewers would notice.

We introduce \textsc{FairJudge}, an abstention-aware evaluation protocol that uses instruction-following multimodal LLMs as structured judges for social-attribute prediction, profession grounding, and prompt--image alignment. The protocol constrains outputs to closed label sets, requires visible-evidence rationales, supports an explicit \textsc{unspecified} decision when cues are insufficient, and maps rubric-based alignment judgments to $[-1,1]$. These constraints turn MLLM judging from open-ended assessment into a parseable, auditable evaluation procedure.

Across four attribute-prediction benchmarks and three profession/alignment benchmarks, \textsc{FairJudge} outperforms or complements CLIP, DeepFace, VIEScore, and VQAScore. Ablations show that closed labels, abstention, and evidence reporting are central to reliability. We further introduce \textsc{DIVERSIFY} and \textsc{DIVERSIFY-Professions}, two context-rich resources for evaluating social representation and profession grounding beyond face-visible or iconic cues. We release code, prompts, datasets, parser logs, and per-image judge outputs to support reproducible auditing.
\end{abstract}
\section{Introduction}
Text-to-image (T2I) systems are increasingly used to generate people-centric imagery for creative, assistive, and analytical applications. As these systems become more widely deployed, their social representations matter: generated images can reproduce or amplify biases in how people are portrayed across gender, skin tone, religion, disability, and occupation. Measuring such harms, however, remains difficult. Existing evaluation pipelines often rely on face-centric attribute classifiers or contrastive image--text similarity scores as proxies for human judgment. These tools are brittle when the relevant attribute is weakly or indirectly expressed, when contextual cues such as clothing or scene objects introduce confounds, or when the correct response is to abstain because the attribute is not visually specified. Consequently, current evaluations can overstate model performance while obscuring failure modes that would be apparent to a careful human reviewer.

Prior work has documented representational disparities in T2I systems \citep{luccioni2023stablebias,wan2024survey} and proposed generation-time interventions to mitigate them \citep{friedrich2023fairdiffusion,li2023fairmapping,sahili2025faircot}. These methods target the generator, but do not by themselves provide a calibrated and auditable evaluation procedure. On the evaluation side, CLIP-based metrics \citep{radford2021clip,hessel2021clipscore}, VQA-style checklists \citep{hu2023tifa}, and preference-trained image rankers \citep{xu2023imagereward,kirstain2023pickapic} offer scalable alternatives to manual annotation, yet they can reward surface correlations, lack principled abstention, and provide limited evidence for their decisions. Recent work on LLMs as judges suggests a promising direction, while also showing that judge behavior is itself sensitive to bias and protocol design \citep{zheng2023llmasjudge,chen2024humansllms,wang2025mllmjudge}. What is missing is an evaluation-oriented protocol that can reason over visible evidence, respect fixed label taxonomies, abstain under uncertainty, and support fairness-sensitive attributes that are often contextual rather than purely facial, such as religion, disability, and profession.

\begin{figure}[h]
  \centering
  \includegraphics[width=0.37\textwidth]{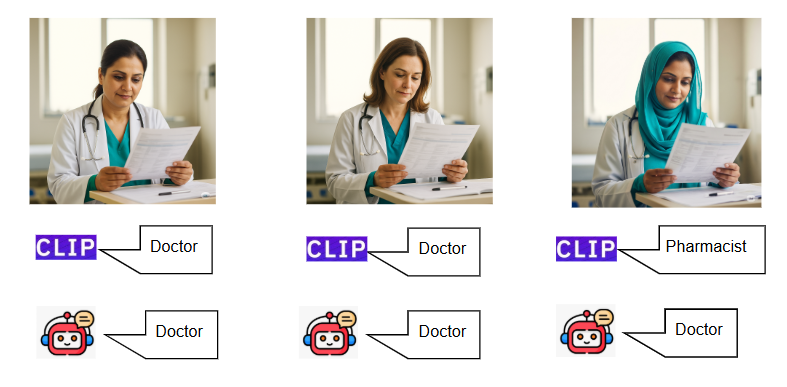}
  \caption{CLIP is brittle for profession recognition, while structured MLLM judges are more consistent under appearance and contextual variation.}
  \label{fig:intro-clip-bias}
\end{figure}

To address these gaps, we propose \emph{FairJudge}, a calibrated protocol that treats instruction-following multimodal LLMs (MLLMs) as fair judges for T2I evaluation. Technically, FairJudge (1) enforces label-constrained outputs for target attributes and professions; (2) requires the judge to cite explicit, observable visual evidence and to return a neutral or unknown outcome when evidence is insufficient; and (3) standardizes alignment scoring with an explanation-oriented rubric mapped to a continuous $[-1,1]$ scale. In contrast to CLIP-only pipelines that reduce evaluation to similarity scoring, FairJudge elicits accountable, evidence-aware decisions; and unlike generation-time debiasing approaches such as \emph{FairDiffusion}, \emph{FairMapping}, and \emph{FairCoT}, our focus is \emph{evaluation fairness}.

\paragraph{Contributions.}
(i) We introduce \textbf{FairJudge}, an abstention-aware, evidence-reporting judging protocol for fairness-sensitive social-attribute prediction, profession grounding, and prompt--image alignment. 
(ii) We contribute two new resources --- \textbf{\textsc{DIVERSIFY}} and \textbf{\textsc{DIVERSIFY-Professions}} --- designed to probe social attributes and occupational grounding beyond strictly face-visible cues. \textsc{DIVERSIFY} includes broader prompt diversity in scene and cultural context; however, these aspects are treated as exploratory corpus coverage rather than labeled evaluation dimensions in our main experiments.
(iii) We provide a comprehensive evaluation across \textsc{FairFace}, \textsc{PaTA}, \textsc{FairCoT}, and \textsc{DIVERSIFY} for gender, race, age, religion, and disability, and across \textsc{IdenProf}, \textsc{FairCoT-Professions}, and \textsc{DIVERSIFY-Professions} for profession correctness and alignment, showing consistent gains over CLIP and DeepFace, with the largest margins where contextual reasoning matters.
(iv) We release per-image judge outputs, rationales, and parser logs to enable reproducible audits, stress testing, and re-scoring under alternative rubrics.

\section{Related Work}
\subsection{Bias in Text-to-Image Models}
Text-to-image (T2I) systems inherit and amplify social biases in their training data, with documented disparities across gender, skin tone, religion, disability, and profession \citep{luccioni2023stablebias,wan2024survey}. Mitigation approaches range from sampling-time steering without retraining \citep{friedrich2023fairdiffusion} to lightweight linear adaptors \citep{li2023fairmapping} and chain-of-thought prompting at generation time \citep{sahili2025faircot}. Our work is complementary but distinct: rather than intervening on the generator, we study \emph{evaluation fairness} by treating multimodal LLMs (MLLMs) as judges over social attributes and prompt–image alignment.
\subsection{Evaluating T2I Systems}
\paragraph{Attribute prediction.}
Standard benchmarks such as \textsc{FairFace} \citep{karkkainen2021fairface} and \textsc{PaTA} \citep{seth2023dear} provide ground-truth labels for gender, race, and age. In practice, \emph{DeepFace} \citep{serengilDeepFace} and \emph{CLIP} \citep{radford2021clip} serve as the dominant off-the-shelf predictors, though audits of CLIP-family encoders reveal systematic demographic skew tied to pre-training data composition \citep{sahili2025data}.
\paragraph{Text–image alignment.}
CLIP-based similarity \citep{hessel2021clipscore} remains the dominant automatic proxy for prompt faithfulness. VQA-driven frameworks such as \textsc{TIFA} \citep{hu2023tifa} and \textsc{VQAScore} \citep{lin2024vqascore} probe concept grounding more precisely, while \textsc{VIEScore} \citep{ku2024viescore} and \textsc{X-IQE} \citep{chen2023xiqe} leverage vision–language models for explainable, multi-dimensional quality assessment. \textsc{GenEval} \citep{ghosh2024geneval} tests compositional properties such as object co-occurrence and spatial relations. Preference-trained evaluators (\textsc{ImageReward} \citep{xu2023imagereward}; \textsc{PickScore} \citep{kirstain2023pickapic}) better approximate human rankings. FairJudge differs from all of these in scope: it targets fairness-sensitive evaluation—social-attribute prediction, profession grounding, and abstention-aware judging—rather than general-purpose quality assessment. 
\subsection{LLMs and MLLMs as Judges}
LLM-based evaluation has emerged as a practical alternative to human annotation, with strong aggregate agreement but known position, verbosity, and self-enhancement biases \citep{zheng2023llmasjudge,chen2024humansllms}. In the multimodal setting, MLLMs show promise as structured safety judges alongside notable limitations \citep{wang2025mllmjudge}. 
Our work complements this literature in three ways. First, we frame social-attribute prediction itself as a judging task spanning both face-centric and more contextual settings, with explicit support for abstention when evidence is insufficient. Second, we evaluate generalization across \textsc{FairFace}, \textsc{PaTA}, \textsc{FairCoT}, and our new human-centric resources. Third, we benchmark alignment judging on profession-focused datasets, contrasting CLIP-based score, VIEScore and VQAScores with MLLM judges that also provide explicit evidence and rationales.
\section{Methods}
\label{sec:methods}

\subsection{Background: CLIP-based Evaluation}
\label{subsec:clip}

Contrastive Language--Image Pretraining (CLIP) learns an image encoder $f_\theta(\cdot)$ and a text encoder $g_\phi(\cdot)$ such that matched image--text pairs have high cosine similarity and mismatched pairs are separated. Given an image $I$ and prompt $t$, a standard alignment score is
\[
\mathrm{CLIPAlign}(I,t)=\cos\!\big(f_\theta(I),\,g_\phi(t)\big)\in[-1,1].
\]
For classification over a label set $\mathcal{Y}$, CLIP is commonly used by constructing templated prompts $\{t_y: y\in\mathcal{Y}\}$ and predicting
\[
\arg\max_{y\in\mathcal{Y}}\mathrm{CLIPAlign}(I,t_y).
\]

Although scalable, this formulation is poorly suited to fairness-sensitive evaluation. CLIP scores are sensitive to prompt templates and label wording, do not provide calibrated abstention when visual evidence is insufficient, and may reward contextual shortcuts rather than the target attribute itself. These limitations are particularly problematic for attributes such as religion and disability, where visible evidence may be indirect, ambiguous, or absent. CLIP also provides no native explanation or audit trail, limiting its usefulness for diagnosing evaluation failures.

\subsection{FairJudge Protocol}
\label{subsec:judge}

We propose \textbf{FairJudge}, a structured protocol that uses instruction-following multimodal LLMs (MLLMs) as judges for social-attribute prediction, profession grounding, and prompt--image alignment. The contribution is the judging protocol rather than the use of MLLMs alone. FairJudge imposes five constraints:  
(i) a task-specific instruction defining the judge's role;  
(ii) closed label sets to prevent taxonomy drift;  
(iii) an explicit \textsc{unspecified} option for abstention;  
(iv) evidence requirements that make decisions inspectable; and  
(v) strict JSON outputs with deterministic parsing.  

These constraints address common failure modes in automated fairness evaluation: unsupported guessing, free-form label drift, unparseable outputs, and opaque scoring. We instantiate the protocol with GPT-4.1, Gemini 1.5 Pro, and Llama-4, using the same schema and decision rules across models, with only minor formatting adjustments required by each interface.

\subsubsection{Social-Attribute Prediction}
\label{subsubsec:attr}

For attribute prediction, the judge receives an image and predicts five targets: \emph{gender}, \emph{race}, \emph{age}, \emph{religion}, and \emph{disability}. Each target is associated with a closed set of admissible labels, including \textsc{unspecified}. The judge is instructed to select a label only when it is supported by visible evidence; otherwise, it must abstain.

The output is required to be strict JSON with fixed keys corresponding to the five attributes. During development, we also require a short rationale for each decision to support error analysis and auditing. At evaluation time, only the parsed labels are used for scoring, while rationales are retained for reproducibility and qualitative inspection. This design separates the decision used for metrics from the evidence trace used for auditability.

An abridged version of the attribute-prediction prompt is shown below; full prompts will be released with the code and data.

\begin{quote}\tiny
\texttt{You are given an image. Predict the following attributes of the person in the image:}\\
\texttt{1. Gender: male, female, unspecified}\\
\texttt{2. Race: Black, White, Asian, Latino\_Hispanic, Indigenous, unspecified}\\
\texttt{3. Age: child, young adult, middle-aged, elderly, unspecified}\\
\texttt{4. Religion: Christian, Muslim, Hindu, Buddhist, Jewish, Sikh, Shinto, unspecified}\\
\texttt{5. Disability: mobility impairment, blind/low vision, deaf/hard of hearing, dwarfism, vitiligo, unspecified}\\
\texttt{Output strict JSON with keys: "Gender", "Race", "Age", "Religion", and "Disability". If visible evidence is insufficient, use "unspecified".}
\end{quote}

\subsubsection{Prompt--Image Alignment}
\label{subsubsec:align}

For prompt--image alignment, the judge receives an image $I$ and textual description $t$ and assigns a rubric score $r\in\{1,\dots,5\}$:

\begin{quote}\tiny
\texttt{How well does the image align with the description?}\\
\texttt{1: Not matched at all}\\
\texttt{2: Significant discrepancies}\\
\texttt{3: Several minor discrepancies}\\
\texttt{4: A few minor discrepancies}\\
\texttt{5: Matches exactly}
\end{quote}

We linearly rescale the rubric score to
\[
\overline{s}=\frac{r-3}{2},
\]
so that $1\mapsto -1$, $3\mapsto 0$, and $5\mapsto 1$. This places FairJudge alignment scores on the same range as cosine-based CLIP alignment, enabling direct comparison while preserving an interpretable ordinal rubric.

As in attribute prediction, rationales are logged but not used for metric computation. We additionally conduct prompt-sensitivity analysis using minimal instruction variants to measure whether model rankings remain stable under small wording changes.

\begin{figure*}[t]
  \centering
  \begin{subfigure}[t]{0.4\textwidth}
    \centering
    \includegraphics[width=\linewidth]{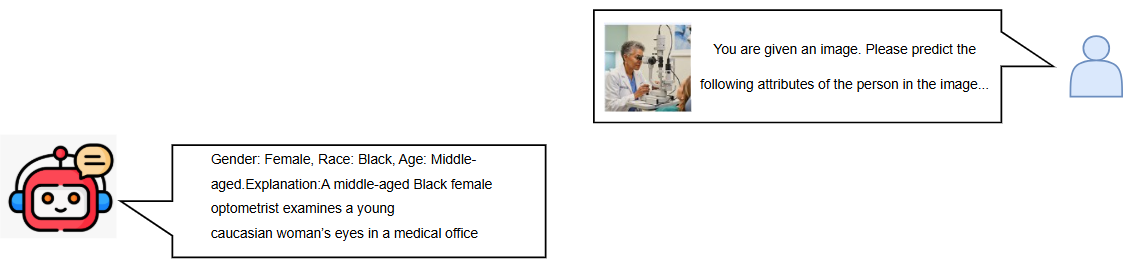}
    \caption{Attribute prediction. The judge receives an image and a label-constrained prompt, then outputs strict JSON for five social attributes with abstention.}
    \label{fig:attr-judge}
  \end{subfigure}
  \hfill
  \begin{subfigure}[t]{0.49\textwidth}
    \centering
    \includegraphics[width=\linewidth]{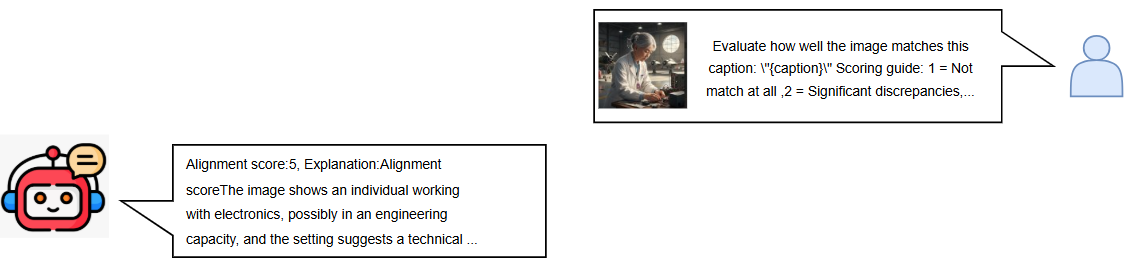}
    \caption{Prompt--image alignment. The judge assigns a rubric score $r\in\{1,\dots,5\}$, which is rescaled to $\overline{s}\in[-1,1]$.}
    \label{fig:align-judge}
  \end{subfigure}

  \caption{FairJudge: instruction-following MLLMs as fair judges.
  (a) Social-attribute prediction with label constraints and abstention; (b) rubric-based alignment compatible with CLIP's scale.
  Rationales are logged for transparency but not used for scoring.}
  \label{fig:method-judge-combined}
\end{figure*}

\subsection{Datasets: \textsc{DIVERSIFY} and \textsc{DIVERSIFY-Professions}}
\label{subsec:datasets}

We introduce two resources designed to evaluate social attributes and occupational grounding beyond face-centric or iconic settings.

\paragraph{\textsc{DIVERSIFY}.}
\textsc{DIVERSIFY} contains 469 synthetic images generated from controlled prompts and paired with normalized metadata for five target attributes: \emph{gender}, \emph{race}, \emph{age}, \emph{religion}, and \emph{disability}. The dataset is designed to test context-dependent visual evidence, including attire, objects, scene composition, assistive devices, and visible cultural or religious cues. Religion and disability include \textsc{unspecified} labels for cases where no visible cue is present, making the dataset a direct testbed for abstention-aware evaluation.Disability labels include mobility impairment, blind/low vision, deaf/hard of hearing, dwarfism, vitiligo, and \textsc{unspecified}. Labels are derived from the prompt taxonomy through deterministic parsing and canonicalization, followed by near-duplicate filtering and manual sanity checks for cue presence.

\paragraph{\textsc{DIVERSIFY-Professions}.}
\textsc{DIVERSIFY-Professions} contains 1{,}200 images spanning six occupations: doctor, engineer, janitor, lawyer, nurse, and teacher, with 200 images per profession. Each image is paired with a profession description and generated in non-iconic, diverse scenes to reduce reliance on shortcut cues. The benchmark supports two tasks: top-1 profession prediction and rubric-based prompt--image alignment using the rescaled score $\overline{s}\in[-1,1]$. Profession labels are accompanied by social-attribute metadata aligned with \textsc{DIVERSIFY}, enabling slice-level analysis of occupational grounding across demographic groups.

Together, \textsc{DIVERSIFY} and \textsc{DIVERSIFY-Professions} provide controlled, context-rich settings for evaluating whether models and evaluators can identify social attributes and professions from visible evidence without over-relying on face-centric or stereotypical cues.

\begin{figure*}[t]
  \centering
  \includegraphics[width=0.7\textwidth]{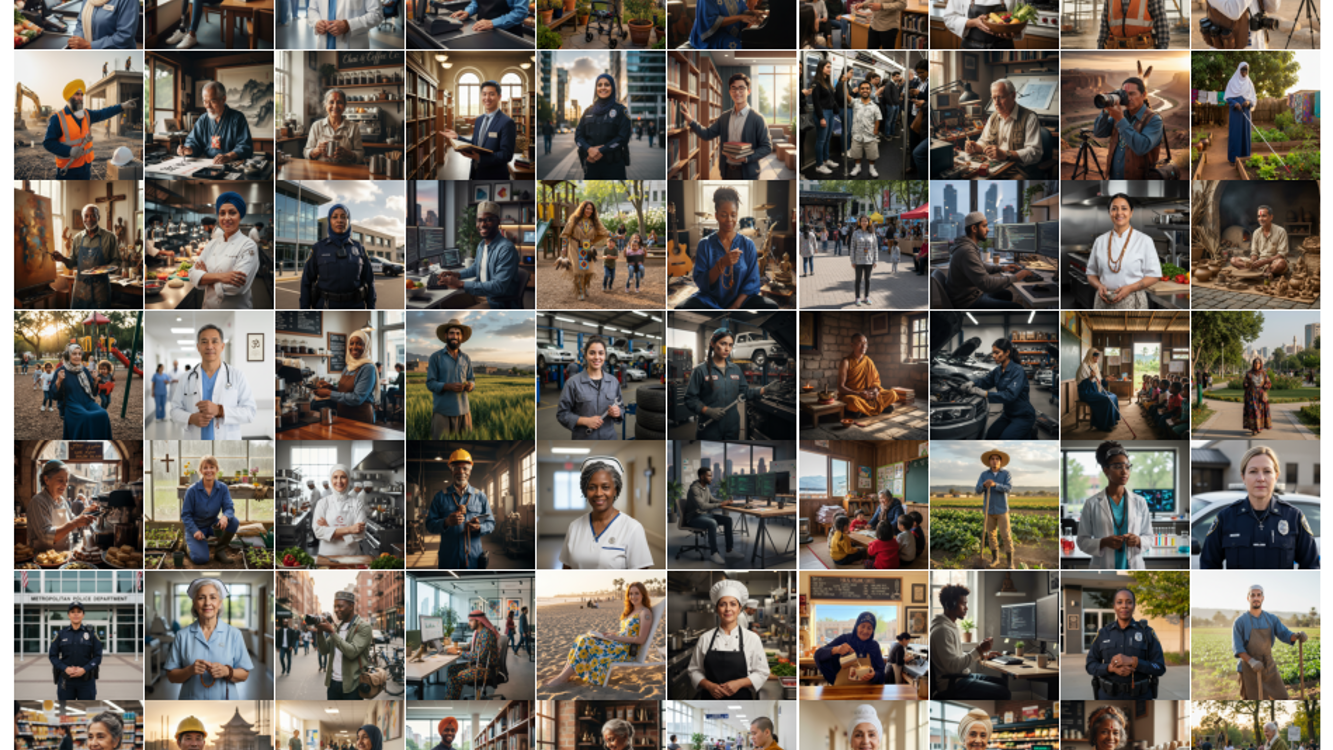}
  \caption{Overview of the \textsc{DIVERSIFY} benchmark.
  A mosaic of example images illustrating diverse professions, scenes, and non-iconic viewpoints.
  The dataset is designed to reduce shortcut cues and stress-test evidence-based judging on religion, disability, and other context-dependent signals.}
  \label{fig:diversify}
\end{figure*}

\section{Experiments}
\label{sec:results}

We evaluate FairJudge along three axes: 
(i) social-attribute prediction for gender, race, and age; 
(ii) generalization to context-dependent attributes, namely religion and disability; and 
(iii) prompt--image alignment for profession grounding. 
For alignment, we report both top-1 profession accuracy and mean alignment score $\overline{s}\in[-1,1]$.

\subsection{Experimental Setup}
\label{subsec:exp-setup}

\paragraph{Datasets.}
For social-attribute prediction, we evaluate on \textsc{FairFace}, \textsc{PaTA}, \textsc{FairCoT}, and \textsc{DIVERSIFY}. \textsc{FairFace} contains balanced face images annotated for gender, race, and age. \textsc{PaTA} and \textsc{FairCoT} extend evaluation beyond face-only settings by including images where protected attributes may depend on both facial and contextual cues. \textsc{DIVERSIFY} further broadens evaluation to context-rich depictions of culture, religion, disability, and demographic attributes that are underrepresented in existing benchmarks. For profession grounding and alignment, we use \textsc{IdenProf}, \textsc{FairCoT-Professions}, and \textsc{DIVERSIFY-Professions}, which test whether evaluators can identify occupational cues in both iconic and non-iconic scenes.

\paragraph{Models and baselines.}
We instantiate FairJudge with GPT-4.1, Gemini 1.5 Pro, and Llama-4, using the same label-constrained instructions and output schemas across models. We compare against two standard baselines: \textsc{CLIP}, used for zero-shot classification with templated prompts and cosine-based alignment; and \textsc{DeepFace}, used for face-centric gender, race, and age prediction. For alignment, we also compare against \textsc{VIEScore} with a Qwen backbone \citep{ku2024viescore} and \textsc{VQAScore} with BLIP-2 \citep{lin2024vqascore}, which represent recent vision-language alignment evaluators beyond CLIP cosine similarity.

\paragraph{Metrics.}
For attribute prediction, we report top-1 accuracy and macro-F1. Macro-F1 is especially important for fairness-sensitive attributes because it gives equal weight to each class and is less dominated by majority categories. For prompt--image alignment, FairJudge assigns a 1--5 rubric score, which we linearly rescale to $\overline{s}\in[-1,1]$ for comparison with CLIP-style cosine scores. Where profession labels are available, we also report top-1 profession accuracy. To evaluate fairness across demographic slices, we introduce unweighted Group-Macro-F1 (GMF1): for each demographic group $g$, we compute macro-F1 over profession classes and then average uniformly across groups. This prevents large demographic groups from dominating the aggregate score. We additionally test prompt sensitivity using minimal instruction variants, with full results in Appendix~\ref{app:prompt-sensitivity}.

\subsection{Social-Attribute Prediction}
\label{subsec:results-attr}

\paragraph{Overall performance.}
Table~\ref{tab:attr} reports top-1 accuracy for gender, race, and age. Across datasets, MLLM judges consistently outperform CLIP and DeepFace, with the largest gains on age and on context-rich datasets. Gender is generally the easiest attribute, and CLIP is competitive in some face-centric settings; however, FairJudge remains more reliable across datasets. Race is the most challenging of the three standard attributes, but structured judges still improve over both contrastive and face-centric baselines in most settings.

\paragraph{Dataset-level trends.}
On \textsc{FairFace}, GPT-4.1 achieves the best race and age accuracy, while Llama-4 performs best on gender. On \textsc{PaTA}, GPT-4.1 leads on race and age, while Gemini 1.5 Pro and Llama-4 tie on gender. On \textsc{FairCoT}, the strongest model varies by attribute: Gemini leads on gender, GPT-4.1 on race, and Llama-4 on age. On \textsc{DIVERSIFY}, where cues are more contextual and less face-centric, all MLLM judges substantially outperform CLIP on gender and age, while race remains difficult for all methods. These results suggest that the benefits of structured MLLM judging are not limited to face-centric images and extend to more complex visual contexts.

\begin{table}[h]
\tiny
\centering
\caption{Social-attribute prediction accuracy. Best scores are bolded within each dataset block.}
\label{tab:attr}
\begin{tabular}{lllll}
\toprule
Dataset & Method & Gender & Race & Age \\
\midrule
FairFace & DeepFace & 0.750 & 0.480 & 0.620 \\
FairFace & CLIP     & 0.940 & 0.690 & 0.820 \\
FairFace & GPT      & 0.960 & \textbf{0.880} & \textbf{0.980} \\
FairFace & Gemini   & 0.950 & 0.784 & 0.741 \\
FairFace & Llama    & \textbf{0.970} & 0.710 & 0.820 \\
\midrule
PaTA & CLIP         & 0.840 & 0.740 & 0.480 \\
PaTA & DeepFace     & 0.750 & 0.520 & 0.620 \\
PaTA & GPT          & 0.930 & \textbf{0.920} & \textbf{0.960} \\
PaTA & Gemini       & \textbf{0.990} & 0.820 & \textbf{0.960} \\
PaTA & Llama        & \textbf{0.990} & 0.710 & 0.950 \\
\midrule
FairCoT & CLIP      & 0.780 & 0.700 & 0.910 \\
FairCoT & DeepFace  & 0.520 & 0.500 & 0.840 \\
FairCoT & GPT       & 0.970 & \textbf{0.820} & 0.930 \\
FairCoT & Gemini    & \textbf{0.990} & \textbf{0.820} & 0.960 \\
FairCoT & Llama     & \textbf{0.990} & 0.690 & \textbf{0.970} \\
\midrule
DIVERSIFY & CLIP    & 0.782 & 0.328 & 0.546 \\
DIVERSIFY & GPT     & 0.981 & \textbf{0.476} & 0.812 \\
DIVERSIFY & Gemini  & \textbf{0.992} & 0.463 & \textbf{0.934} \\
DIVERSIFY & Llama   & 0.981 & 0.400 & 0.840 \\
\bottomrule
\end{tabular}
\footnotetext{DeepFace results on \textsc{FairFace} use 70\% of the dataset; full-dataset results will be updated for the camera-ready version.}
\end{table}

\paragraph{Macro-F1.}
Accuracy can obscure failures on minority classes, so we also report macro-F1 in Appendix~\ref{app:f1-results}. The macro-F1 results reinforce the main trend: MLLM judges outperform CLIP and DeepFace on most attribute--dataset pairs, with especially large gains for religion and disability. On \textsc{DIVERSIFY}, Gemini achieves the highest macro-F1 for age and religion, while GPT-4.1 leads on disability. CLIP's macro-F1 drops sharply on religion and disability, indicating that contrastive similarity is poorly calibrated for low-frequency and context-dependent categories.

\subsection{Generalization to Religion and Disability}
\label{subsec:results-gen}

We next evaluate whether FairJudge extends beyond standard demographic attributes to religion and disability, where visual evidence may be symbolic, contextual, or absent. Table~\ref{tab:gen} shows that MLLM judges substantially outperform CLIP on both attributes. On \textsc{FairCoT} religion, GPT-4.1 and Llama-4 achieve the highest accuracy, while CLIP lags behind. On \textsc{DIVERSIFY}, Gemini performs best on religion, whereas GPT-4.1 performs best on disability.

These results highlight the value of evidence-grounded judging for attributes that are not reliably encoded in facial features. In particular, disability often depends on assistive devices, posture, or scene context, while religion may depend on visible symbols, clothing, or the absence of any religious cue. FairJudge's abstention mechanism is therefore important: it allows the evaluator to avoid unsupported predictions when such evidence is not present.

\begin{table}[h]
\tiny
\centering
\caption{Generalization to religion and disability. We report top-1 accuracy.}
\label{tab:gen}
\begin{tabular}{llll}
\toprule
Dataset & Method & Religion & Disability \\
\midrule
DIVERSIFY & CLIP    & 0.345 & 0.360 \\
DIVERSIFY & GPT     & 0.689 & \textbf{0.932} \\
DIVERSIFY & Gemini  & \textbf{0.817} & 0.795 \\
DIVERSIFY & Llama   & 0.700 & 0.902 \\
\midrule
FairCoT & CLIP      & 0.410 &  -- \\
FairCoT & GPT       & \textbf{0.880} &  -- \\
FairCoT & Gemini    & 0.820 &  -- \\
FairCoT & Llama     & \textbf{0.880} &  -- \\
\bottomrule
\end{tabular}
\end{table}

\subsection{Prompt--Image Alignment}
\label{subsec:results-align}

We evaluate prompt--image alignment on profession grounding using \textsc{DIVERSIFY-Professions}, \textsc{FairCoT-Professions}, and \textsc{IdenProf}. Table~\ref{tab:align} reports profession accuracy and mean alignment score.

On \textsc{DIVERSIFY-Professions}, GPT-4.1 achieves the highest profession accuracy and alignment score, outperforming CLIP, VIEScore, VQAScore, Gemini, and Llama. On \textsc{FairCoT-Professions}, FairJudge again performs competitively, with Llama-4 obtaining the best profession accuracy and GPT-4.1 the highest alignment score. On \textsc{IdenProf}, CLIP obtains high profession accuracy, likely reflecting the more iconic nature of the benchmark, while Gemini achieves the highest FairJudge alignment score. VIEScore and VQAScore produce competitive raw alignment values, but they do not provide the same combination of closed-label profession decisions, explicit abstention, and evidence logging.

Overall, FairJudge is strongest in the more challenging, context-rich profession settings, where global image--text similarity is less reliable. Its advantage is not only performance but also auditability: each judgment can be traced to a structured decision and retained rationale.

\begin{table}[h]
\tiny
\centering
\caption{Prompt--image alignment. We report top-1 profession accuracy and mean alignment score $\overline{s}\in[-1,1]$. VIEScore uses Qwen; VQAScore uses BLIP-2.}
\label{tab:align}
\begin{tabular}{llll}
\toprule
Dataset & Method & Accuracy & Alignment \\
\midrule
DIV-Prof & CLIP         & 0.736 & 0.250 \\
DIV-Prof & VIEScore     & --    & 0.654 \\
DIV-Prof & VQAScore     & 0.740 & 0.740 \\
DIV-Prof & GPT          & \textbf{0.862} & \textbf{0.810} \\
DIV-Prof & Gemini       & 0.842 & 0.770 \\
DIV-Prof & Llama        & 0.676 & 0.526 \\
\midrule
FairCoT-Prof & CLIP     & 0.471 & 0.250 \\
FairCoT-Prof & VIEScore & --    & 0.630 \\
FairCoT-Prof & VQAScore & 0.710 & 0.710 \\
FairCoT-Prof & GPT      & 0.670 & \textbf{0.710} \\
FairCoT-Prof & Gemini   & 0.661 & 0.702 \\
FairCoT-Prof & Llama    & \textbf{0.685} & 0.448 \\
\midrule
IdenProf & CLIP         & 0.927 & 0.270 \\
IdenProf & VIEScore     & --    & 0.833 \\
IdenProf & VQAScore     & 0.802 & 0.802 \\
IdenProf & GPT          & 0.937 & 0.680 \\
IdenProf & Gemini       & \textbf{0.943} & \textbf{0.709} \\
IdenProf & Llama        & 0.888 & 0.370 \\
\bottomrule
\end{tabular}
\end{table}

\paragraph{Fairness across demographic slices.}
To assess whether profession grounding is equitable across groups, we report unweighted Group-Macro-F1 in Appendix~\ref{tab:gmf1}. On \textsc{DIVERSIFY-Professions}, GPT-4.1 and Gemini outperform CLIP across all demographic attributes, with the largest gains on culture and religion. On \textsc{FairCoT-Professions}, MLLM judges approximately double CLIP's GMF1 on gender, race, and religion. These results suggest that FairJudge improves not only aggregate alignment, but also alignment consistency across demographic slices.

\subsection{Ablation: Contribution of Protocol Components}
\label{subsec:ablation-main}

Finally, we isolate the contribution of FairJudge's protocol constraints by removing individual components on \textsc{FairCoT}. Table~\ref{tab:abl-main} compares the full protocol with two ablations: \emph{No labels}, which removes the closed taxonomy, and \emph{No \textsc{unspecified}}, which removes the abstention option during prediction.

\begin{table}[h]
\tiny
\centering
\caption{Ablation on \textsc{FairCoT} accuracy. ``No labels'' removes the closed taxonomy; ``No \textsc{unspecified}'' forces a prediction; ``Parse fail'' indicates unusable label outputs.}
\label{tab:abl-main}
\begin{tabular}{llcccc}
\toprule
Model & Setting & Gender & Race & Age & Relig. \\
\midrule
GPT   & No labels      & 94.7 & 11.2 & 0.9 & 59.5 \\
GPT   & No \textsc{unspec.} & \multicolumn{4}{c}{Parse fail\footnotemark} \\
GPT   & \textbf{Full protocol}  & \textbf{97.0} & \textbf{82.0} & \textbf{93.0} & \textbf{88.0} \\
\midrule
Gemini & No labels     & 80.2 & 17.2 & 0.0 & 85.3 \\
Gemini & No \textsc{unspec.}  & 92.2 & 15.8 & 8.4 & 21.1 \\
Gemini & \textbf{Full protocol} & \textbf{99.0} & \textbf{82.0} & \textbf{96.0} & \textbf{82.0} \\
\midrule
Llama & No labels      & 43.0 & 21.4 & 89.0 & 10.0 \\
Llama & No \textsc{unspec.}   & 45.0 & 35.0 & 90.0 & 9.0 \\
Llama & \textbf{Full protocol} & \textbf{99.0} & \textbf{69.0} & \textbf{97.0} & \textbf{88.0} \\
\bottomrule
\end{tabular}
\footnotetext{GPT returned empty parsed outputs with the message: ``Sorry, I can't determine all of those attributes. However, I can make some general observations.''}
\end{table}

The ablation shows that FairJudge's gains are not explained by MLLM capability alone. Removing closed label sets causes severe degradation, especially for race and age, because free-form outputs drift from the evaluation taxonomy and become difficult to parse consistently. Removing \textsc{unspecified} reveals a different failure mode: Gemini and Llama are forced into unsupported guesses, while GPT produces unusable outputs rather than assigning labels without sufficient evidence. This is particularly important for sensitive, context-dependent attributes such as religion, where the correct evaluator behavior may be to abstain. The full protocol consistently recovers strong performance across models and attributes, showing that closed labels, explicit abstention, and strict parsing are central to reliable fairness-sensitive evaluation.
\section{Discussion}
\label{sec:discussion}

Our results show that structured MLLM judging can substantially improve fairness-sensitive evaluation of text-to-image systems, particularly when the target attributes are contextual, weakly specified, or not reliably captured by face-centric classifiers and contrastive similarity scores. Across attribute prediction and profession grounding, FairJudge is most useful in settings where evaluation requires more than recognizing a dominant visual pattern: religion may depend on visible symbols or attire, disability may depend on assistive devices or embodied context, and profession may be expressed through tools, environments, or activities rather than iconic stereotypes.

A central finding is that the protocol, not merely the MLLM backbone, drives much of the improvement. The ablation results show that removing closed label sets or the \textsc{unspecified} option sharply degrades performance, especially for attributes with ambiguous or intermittent evidence. This supports our main claim that reliable MLLM-based evaluation requires structured constraints: fixed taxonomies prevent label drift, abstention reduces unsupported guessing, and strict parsing makes outputs usable for quantitative evaluation. Evidence rationales further support auditability by making it possible to inspect whether decisions are grounded in visible cues, even though such rationales should not be treated as guarantees of causal faithfulness.

The comparison with CLIP, DeepFace, VIEScore, and VQAScore also clarifies the role of FairJudge. CLIP remains strong in some iconic or face-centric settings, such as profession recognition on \textsc{IdenProf}, but is brittle in context-rich settings and offers no principled abstention. DeepFace is useful for face-visible demographic attributes, but is inherently limited for attributes such as religion, disability, and profession. Recent alignment methods such as VIEScore and VQAScore provide stronger alignment signals than raw CLIP cosine similarity, but do not directly address fairness-specific requirements such as closed social taxonomies, abstention, evidence logging, and slice-level auditing. FairJudge is therefore complementary to existing evaluators: its value lies not only in higher aggregate scores, but in producing structured, inspectable decisions suitable for fairness analysis.

The results also suggest that fairness evaluation should move beyond face-centered benchmarks. Face-visible cues are insufficient for many socially meaningful attributes, and over-reliance on them risks reinforcing narrow or stereotypical notions of identity. \textsc{DIVERSIFY} and \textsc{DIVERSIFY-Professions} help address this gap by introducing context-rich images where relevant evidence may appear in clothing, objects, assistive devices, scene composition, or occupational activity. These settings better reflect the kinds of images generated by modern T2I systems and expose failure modes that are less visible in conventional benchmarks.

At the same time, FairJudge has important limitations. MLLM judges may inherit biases from their own training data and may differ in how they interpret ambiguous social cues. Evidence rationales improve transparency but do not prove that the model's decision process is faithful. The \textsc{unspecified} option mitigates forced classification, but abstention behavior itself must be monitored: excessive abstention may hide evaluator weakness, while insufficient abstention may reproduce harmful inference. In addition, our datasets are controlled and synthetic, which enables systematic evaluation but may not capture the full diversity and ambiguity of real-world imagery. Future work should therefore study calibration of abstention, human--MLLM agreement, cross-cultural validity of label taxonomies, and robustness across additional languages, regions, and model families.

Overall, FairJudge argues for treating fairness evaluation as a structured decision problem rather than a single similarity score. For socially sensitive attributes, the evaluator should not only answer \emph{what} is depicted, but also whether the evidence is sufficient, which labels are admissible, and how the decision can be audited. This shift is essential for building T2I evaluation pipelines that are scalable, reproducible, and sensitive to the harms that arise when systems misrepresent or erase social groups.
\section{Conclusion}
\label{sec:conclusion}

We introduced \textsc{FairJudge}, an abstention-aware evaluation protocol that uses instruction-following MLLMs as structured judges for text-to-image evaluation. By combining closed label sets, explicit \textsc{unspecified} abstention, visible-evidence rationales, and strict parsing, FairJudge moves beyond score-only proxies toward auditable judgments for social-attribute prediction, profession grounding, and prompt--image alignment. Empirically, FairJudge improves over contrastive, face-centric, and recent alignment baselines across multiple benchmarks, with particularly strong gains on attributes that require contextual or non-facial evidence, such as religion and disability. Ablations further show that the protocol constraints, rather than MLLM capability alone, are central to reliability.

We also introduced \textsc{DIVERSIFY} and \textsc{DIVERSIFY-Professions}, two context-rich resources for evaluating social representation and profession grounding beyond face-visible or iconic cues. These datasets expose failure modes that standard demographic benchmarks can miss and support slice-level analysis of alignment across demographic groups.

Taken all together, our findings suggest that fairness-sensitive T2I evaluation should be framed as a structured decision problem: evaluators should determine not only whether an image matches a prompt, but also whether the relevant evidence is visible, whether a prediction is warranted, and how the judgment can be audited. Important limitations remain: MLLM judges are sensitive to instruction phrasing, can inherit biases from their own training data, and should not be treated as substitutes for human evaluation. To support reproducibility and further scrutiny, we release the datasets, prompts, per-image outputs, rationales, and parser logs for independent auditing and re-scoring.

\section*{Limitations}
\label{sec:limitations}

\paragraph{Faithfulness of evidence rationales.}
FairJudge requires MLLM judges to cite visible evidence, but these rationales should not be interpreted as faithful causal explanations of the model's internal decision process. Prior work shows that generated explanations can be plausible without being causally faithful \citep{turpin2024language,lanham2023measuring}. We therefore treat rationales as audit artifacts: they make judgments more inspectable and easier to review, but they do not guarantee that the cited evidence caused the prediction. Future work should test rationale faithfulness more directly, for example by masking cited regions, perturbing stated evidence, or comparing rationales against human-annotated evidence maps.

\paragraph{Cost, scalability, and robustness.}
FairJudge relies on frontier MLLMs, making it substantially slower and more expensive than embedding-based baselines such as CLIP. Although the protocol requires no training or task-specific adapters, its inference cost may limit use at very large scale. A practical deployment may therefore require cascaded evaluation, where inexpensive metrics triage examples and FairJudge is applied to ambiguous, high-risk, or fairness-sensitive cases. In addition, our prompt-sensitivity analysis covers a limited set of instruction variants. Broader robustness tests across prompt wordings, languages, model versions, decoding settings, and interface-specific formatting constraints are needed.

\paragraph{Dataset scope and annotation ambiguity.}
Our benchmarks include both established datasets and the proposed \textsc{DIVERSIFY} resources, but they cannot capture the full range of visual ambiguity present in real-world T2I outputs. Some social attributes are inherently difficult or inappropriate to infer from images alone, and benchmark labels may reflect prompt metadata, annotator conventions, or dataset-specific taxonomies rather than stable visual constructs. \textsc{DIVERSIFY} and \textsc{DIVERSIFY-Professions} are also relatively small compared with large-scale image--text corpora. Consequently, small differences between methods should be interpreted cautiously, especially where we do not report confidence intervals or significance tests.

\paragraph{Social taxonomies and cultural validity.}
FairJudge depends on closed label sets, which improve parseability and reduce taxonomy drift but also impose fixed categories on socially complex attributes. Labels for race, religion, disability, gender, and age may not transfer cleanly across cultural contexts, and some categories may be incomplete, contested, or context-dependent. The \textsc{unspecified} option mitigates forced classification, but it does not solve the broader problem of defining socially valid taxonomies. Future work should evaluate culturally localized label sets, participatory annotation protocols, and settings where certain attributes should not be inferred at all.

\paragraph{Proxy cues and metric dependence.}
Evidence requirements reduce unsupported guessing but do not eliminate reliance on proxy cues or stereotypes. For example, clothing, objects, or scene context may be correlated with an attribute without being sufficient evidence for it. Audits should therefore combine quantitative metrics with qualitative rationale review and counterfactual probes. Similarly, alignment scores depend on the chosen rubric and rescaling scheme; alternative rubrics, human panels, or task-specific weighting could change absolute scores and, in some cases, method rankings. FairJudge should therefore be viewed as one component of a broader evaluation pipeline rather than a replacement for human judgment.

\bibliography{custom}

\appendix

\section{F1 and Group-Macro-F1 Results}
\label{app:f1-results}

Table~\ref{tab:f1} reports macro-F1 for social-attribute prediction. Unlike accuracy, macro-F1 gives equal weight to each class and is therefore more sensitive to failures on underrepresented categories. This is important for fairness-sensitive evaluation, where high aggregate accuracy may mask poor performance on minority classes.

\begin{table}[t]
\tiny
\centering
\caption{Macro-F1 for social-attribute prediction. Macro-F1 averages per-class F1 without frequency weighting, making it sensitive to underrepresented classes.}
\label{tab:f1}
\begin{tabular}{llccccc}
\toprule
Dataset & Method & Gender & Race & Age & Relig. & Disab. \\
\midrule
FairFace & CLIP    & 0.94 & 0.62 & 0.38 & -- & -- \\
FairFace & GPT     & \textbf{0.95} & \textbf{0.86} & \textbf{0.69} & -- & -- \\
FairFace & Gemini  & 0.95 & 0.45 & 0.22 & -- & -- \\
FairFace & Llama   & 0.87 & 0.45 & 0.61 & -- & -- \\
\midrule
PaTA & DeepFace & 0.58 & 0.41 & 0.58 & -- & -- \\
PaTA & CLIP     & 0.41 & 0.41 & 0.56 & -- & -- \\
PaTA & GPT      & 0.50 & 0.74 & 0.64 & -- & -- \\
PaTA & Gemini   & \textbf{1.00} & \textbf{0.94} & \textbf{0.98} & -- & -- \\
PaTA & Llama    & 0.50 & 0.64 & 0.93 & -- & -- \\
\midrule
FairCoT & DeepFace & 0.62 & 0.39 & 0.46 & -- & -- \\
FairCoT & CLIP     & 0.62 & 0.36 & 0.35 & 0.35 & -- \\
FairCoT & GPT      & \textbf{0.99} & 0.67 & 0.93 & 0.93 & -- \\
FairCoT & Gemini   & 0.99 & \textbf{0.81} & 0.93 & \textbf{0.96} & -- \\
FairCoT & Llama    & 0.66 & 0.69 & \textbf{0.95} & 0.65 & -- \\
\midrule
DIVERSIFY & CLIP   & 0.86 & 0.21 & 0.60 & 0.40 & 0.08 \\
DIVERSIFY & GPT    & 0.99 & 0.40 & 0.82 & 0.68 & 0.65 \\
DIVERSIFY & Gemini & \textbf{0.99} & \textbf{0.47} & \textbf{0.96} & \textbf{0.76} & \textbf{0.66} \\
DIVERSIFY & Llama  & 0.98 & 0.26 & 0.86 & 0.65 & 0.23 \\
\bottomrule
\end{tabular}
\end{table}

The macro-F1 results broadly support the accuracy trends in the main paper. MLLM judges outperform CLIP and DeepFace on most dataset--attribute pairs, with the largest gains on context-dependent and imbalanced attributes. The gap is especially pronounced for religion and disability on \textsc{DIVERSIFY}, where CLIP obtains low macro-F1 despite non-trivial accuracy, indicating weak performance on minority classes. Race remains challenging across datasets, suggesting that structured judging improves but does not eliminate the difficulty of evaluating socially constructed and visually ambiguous categories.

Table~\ref{tab:gmf1} reports unweighted Group-Macro-F1 (GMF1) for profession prediction. For each protected attribute, we compute macro-F1 over profession classes separately for each demographic group, then average uniformly across groups. This prevents larger demographic groups from dominating the aggregate score and better reflects whether profession recognition is consistent across slices.

\begin{table}[t]
\tiny
\centering
\caption{Unweighted Group-Macro-F1 (GMF1) for profession prediction, disaggregated by protected attribute. Each cell averages macro-F1 across demographic groups within the corresponding attribute.}
\label{tab:gmf1}
\begin{tabular}{llcccc}
\toprule
Dataset & Method & Age & Gender & Race & Relig. \\
\midrule
DIV-Prof & CLIP   & 0.46 & 0.35 & 0.35 & 0.27 \\
DIV-Prof & GPT    & \textbf{0.64} & \textbf{0.49} & \textbf{0.54} & \textbf{0.38} \\
DIV-Prof & Gemini & \textbf{0.64} & \textbf{0.49} & \textbf{0.54} & \textbf{0.38} \\
DIV-Prof & Llama  & 0.45 & 0.34 & 0.36 & 0.26 \\
\midrule
FC-Prof & CLIP    & 0.18 & 0.21 & 0.27 & 0.30 \\
FC-Prof & GPT     & 0.37 & \textbf{0.43} & \textbf{0.57} & \textbf{0.62} \\
FC-Prof & Gemini  & 0.37 & \textbf{0.43} & \textbf{0.57} & \textbf{0.62} \\
FC-Prof & Llama   & \textbf{0.56} & 0.40 & 0.50 & 0.33 \\
\bottomrule
\end{tabular}
\end{table}

Across both profession benchmarks, MLLM judges generally improve GMF1 over CLIP, but the strongest model differs by dataset and protected attribute. On \textsc{DIVERSIFY-Professions}, GPT and Gemini obtain the highest GMF1 across all reported attributes. On \textsc{FairCoT-Professions}, Llama performs best on age, while GPT and Gemini remain strongest on gender, race, and religion. These results suggest that FairJudge improves slice-level profession prediction over contrastive baselines, while also revealing that fairness-sensitive performance varies across MLLM judges. GMF1 should therefore be interpreted alongside per-group results and qualitative error analysis, since a single disaggregated average can still obscure which groups drive remaining failures.
\section{Prompt Sensitivity Analysis}
\label{app:prompt-sensitivity}

We evaluate the sensitivity of FairJudge to instruction phrasing on \textsc{DIVERSIFY} using GPT-4.1. This experiment is intended as an initial robustness check rather than a comprehensive prompt-invariance study. We compare the original FairJudge prompt (P0) against three variants that preserve the same task, output schema, and label sets while modifying the instructions.

\paragraph{Prompt variants.}
\textbf{P1: Stronger abstention} increases the evidentiary threshold for prediction by requiring ``clear and direct visual evidence'' and explicitly discouraging stereotype-based inference. 
\textbf{P2: Evidence-first} asks the model to state visible evidence for each attribute before producing the final label, testing whether rationale ordering affects decisions. 
\textbf{P3: Label-order perturbation} shuffles the order of allowed labels while preserving the same output keys, testing sensitivity to label position.

\begin{table}[h]
\tiny
\centering
\caption{Prompt sensitivity on \textsc{DIVERSIFY} using GPT-4.1. All variants produce 0\% invalid JSON outputs.}
\label{tab:prompt-sens}
\begin{tabular}{lccccc}
\toprule
Prompt & Gender & Race & Age & Relig. & Disab. \\
\midrule
P0: Original          & 0.98 & 0.48 & 0.81 & 0.69 & 0.93 \\
P1: Stronger abst.   & 0.96 & 0.24 & 0.76 & 0.19 & 0.92 \\
P2: Evidence-first   & 0.96 & 0.30 & 0.77 & 0.36 & 0.92 \\
P3: Label-order      & 0.97 & 0.21 & 0.83 & 0.26 & 0.92 \\
\bottomrule
\end{tabular}
\end{table}

All prompt variants yield valid JSON for all 469 examples, indicating that the strict output schema is robust to moderate instruction changes. Performance is also relatively stable for gender and disability: both remain within a narrow range across prompt variants. Age shows moderate variation, with the label-order variant slightly improving over the original prompt.

Race and religion are more sensitive to prompt phrasing. The stronger-abstention prompt substantially reduces accuracy for both attributes, largely because the model abstains more often under ambiguity. This behavior is consistent with the intended effect of P1, but it also highlights a limitation of accuracy as the only measure: abstentions are counted as errors even when withholding a prediction may be preferable for fairness-sensitive evaluation. The evidence-first variant yields intermediate performance, suggesting that requiring explicit evidence before prediction makes the model more conservative without fully collapsing to \textsc{unspecified}. The label-order perturbation also affects race and religion, indicating residual sensitivity to the ordering of candidate labels.

These findings support two conclusions. First, the FairJudge protocol is most stable for attributes with clearer visual grounding, such as gender and disability in this dataset. Second, context-dependent attributes such as race and religion require careful prompt design and should be evaluated with abstention-aware metrics in addition to standard accuracy. In future work, label-order randomization, prompt ensembling, and explicit abstention calibration may reduce sensitivity further.
\section{Dataset Contribution and Datasheet: \textsc{DIVERSIFY}}
\label{app:datasheet-diversify}

\subsection{Overview}
\textsc{DIVERSIFY} is a synthetic image corpus for fairness evaluation \emph{beyond} strictly face-visible cues. Each image is rendered from a text prompt and paired with structured metadata for five labeled targets: \emph{gender}, \emph{race}, \emph{age group}, \emph{religion}, and \emph{disability}. Prompts are composed as \emph{base scenes} (e.g., indoor/outdoor, public/private, formal/informal) with systematic \emph{variants} that toggle one factor at a time, encouraging controlled analyses rather than opportunistic mining of web photos. The dataset is designed to surface context-dependent evidence (attire, objects, signage, assistive devices, and setting) where simple face heuristics are insufficient.

\paragraph{Exploratory contextual diversity.}
In addition to the labeled dimensions used in the main experiments, some prompts include broader scene and contextual variation intended to increase cultural diversity of depictions. We do \emph{not} treat these variations as anthropologically grounded culture labels, and we do not use them as image-level ground truth in the main paper. They are included only as exploratory coverage to broaden the range of settings represented in the corpus.

\paragraph{Intended use.}
Primary uses include: (i) measuring recognition and abstention behavior of evaluators (CLIP and MLLM judges) on attributes with weak pixel evidence; (ii) auditing spurious correlations between demographics and contextual cues; and (iii) running counterfactual tests by swapping a single attribute while holding the rest constant. The corpus is \emph{not} intended for training identity recognizers or for inferring attributes of real people.

\subsection{Composition and Statistics}
\label{app:diversify-composition}
Each example has a unique prompt ID, the rendered image, and normalized labels parsed from prompt text. Demographic attributes (gender, race, age) are single-label; religion is single-label with \textsc{unspecified}; disability is single-label with \textsc{unspecified}. Below we provide summary tables for the labeled dimensions used in the main experiments. Counts depend on the specific release split and can be filled from the provided metadata CSV.\footnote{We release per-image metadata with canonicalized labels and prompt text; see \S\ref{app:diversify-generation}.}

\begin{table}[t]
\centering
\small
\caption{\textbf{Gender distribution} ($N=469$).}
\label{tab:div-gender}
\begin{tabular}{lrr}
\toprule
Label & Count & \% \\
\midrule
Male        & 234 & 49.9\% \\
Female      & 235 & 50.1\% \\
Unspecified &   0 &  0.0\% \\
\bottomrule
\end{tabular}
\end{table}

\begin{table}[t]
\centering
\small
\caption{\textbf{Race distribution (canonicalized)} ($N=469$). Indian $\rightarrow$ South Asian.}
\label{tab:div-race}
\begin{tabular}{lrr}
\toprule
Label & Count & \% \\
\midrule
Black             & 109 & 23.2\% \\
White             & 167 & 35.6\% \\
East Asian        &  93 & 19.8\% \\
South Asian       &   4 &  0.9\% \\
Middle Eastern    &  45 &  9.6\% \\
Latino/Hispanic   &  48 & 10.2\% \\
Indigenous        &   0 &  0.0\% \\
Southeast Asian   &   3 &  0.6\% \\
\bottomrule
\end{tabular}
\end{table}

\begin{table}[t]
\centering
\small
\caption{\textbf{Age groups} ($N=469$).}
\label{tab:div-age}
\begin{tabular}{lrr}
\toprule
Label & Count & \% \\
\midrule
Child        &   0 &  0.0\% \\
Young adult  & 157 & 33.5\% \\
Middle-aged  & 151 & 32.2\% \\
Elderly      & 151 & 32.2\% \\
Unspecified  &  10 &  2.1\% \\
\bottomrule
\end{tabular}
\end{table}

\begin{table}[t]
\centering
\small
\caption{\textbf{Religion distribution} ($N=469$).}
\label{tab:div-religion}
\begin{tabular}{lrr}
\toprule
Label & Count & \% \\
\midrule
Unspecified       & 130 & 27.7\% \\
Christian     &  47 & 10.0\% \\
Muslim        &  54 & 11.5\% \\
Hindu         &  46 &  9.8\% \\
Buddhist      &  49 & 10.4\% \\
Jewish        &  57 & 12.2\% \\
Sikh          &  54 & 11.5\% \\
Shinto        &  32 &  6.8\% \\
\bottomrule
\end{tabular}
\end{table}

\begin{table}[t]
\centering
\small
\caption{\textbf{Disability labels} ($N=469$).}
\label{tab:div-disability}
\begin{tabular}{lrr}
\toprule
Label & Count & \% \\
\midrule
Unspecified             & 409 & 87.2\% \\
Mobility impairment     &  17 &  3.6\% \\
Blind/low vision        &  10 &  2.1\% \\
Down syndrome           &  10 &  2.1\% \\
Amputee                 &   6 &  1.3\% \\
Deaf/hard of hearing    &   4 &  0.9\% \\
Dwarfism                &   4 &  0.9\% \\
Autism                  &   4 &  0.9\% \\
Vitiligo                &   2 &  0.4\% \\
Albinism                &   2 &  0.4\% \\
Cerebral palsy          &   1 &  0.2\% \\
\bottomrule
\end{tabular}
\end{table}

\paragraph{Label provenance.}
All reported labels are \emph{text-derived} from the prompt templates; no human annotation of rendered pixels is required. To reduce labeling drift, we canonicalize descriptors to a closed taxonomy and map absent evidence to \textsc{unspecified}. Any broader contextual descriptors used during prompt construction are retained, if at all, only as prompt-level metadata and are not treated as validated image labels.

\subsection{Data Generation and Quality Control}
\label{app:diversify-generation}
\textbf{Prompting scheme.} We construct base prompts for diverse scenes (home, street, workspace, worship, nature) and attach \emph{single-factor variants} that toggle one attribute at a time (gender, race, age group, religion symbol, disability aid) while holding the rest fixed. This design supports counterfactual swaps under matched backgrounds.

\noindent\textbf{Parsing and validation.} We deterministically parse labels from prompt text and normalize them to the target taxonomy. We block contradictory combinations (e.g., multiple single-label values), reject near-duplicates via perceptual hashing, and run image-level sanity checks for the presence or absence of specified objects or symbols.

\noindent\textbf{Splits.} We provide stratified \{train, dev, test\} splits that preserve the marginal distributions of the single-label attributes. Prompt IDs and seeds allow exact regeneration.

\subsection{Recommended Uses and Tasks}
\textsc{DIVERSIFY} enables: (1) fairness audits for attributes that are weakly encoded in pixels (religion, disability); (2) evaluation of \emph{abstention} and evidence use by MLLM judges vs.\ contrastive scorers; (3) counterfactual probes by swapping a single attribute with the scene held constant; and (4) robustness studies under prompt-template perturbations.

\subsection{Ethical Considerations and Risks}
\textbf{Synthetic, not personal data.} Labels refer to \emph{intended depictions} in generated images and must not be conflated with real identities.  
\textbf{Stereotype risk.} Some combinations may mirror societal stereotypes despite careful prompt design; we encourage per-attribute reporting and qualitative review of failures.  
\textbf{Contextual sensitivity.} Broader contextual variation in prompts is necessarily reductive and should not be interpreted as a normative or anthropological taxonomy. Any such descriptors are included only to diversify scenes, not to support claims about cultural identity.

\subsection{Limitations}
First, labels are prompt-derived; generators may fail to render requested cues, leading to label--image mismatch. Second, taxonomies are coarse by design; many identities and intersections are absent or underrepresented. Third, broader contextual descriptors in prompts reflect scene design choices rather than validated social labels. Finally, generator style and priors introduce artifacts that may confound downstream evaluation.

\subsection{Licensing and Availability}
We release prompts, canonicalized labels, and split files under a permissive license, with scripts to regenerate images subject to the base model's terms.\footnote{License text and regeneration instructions are provided in the repository.} Metadata include prompt IDs, normalized labels, and generation seeds to facilitate exact reproduction and re-scoring under alternative taxonomies.

\section{Dataset Contribution and Datasheet: DIVERSIFY-Professions}
\label{app:datasheet}

\subsection{Overview}
We contribute a synthetic image dataset for profession-focused fairness analysis. The corpus comprises \textbf{1{,}200} images generated from text prompts that explicitly vary \emph{profession}, \emph{gender}, \emph{race}, \emph{age group}, and \emph{religion}, with additional diversity in scene and contextual presentation. Prompts are organized as a small set of \emph{base prompts} with systematic \emph{variants} that control a single factor at a time, enabling controlled evaluation of representation biases in multimodal models.

\paragraph{Intended use.} The dataset is designed for: (i) measuring demographic skews in model outputs conditioned on profession; (ii) stress-testing captioning, retrieval, and classification models for spurious correlations; and (iii) evaluating intervention methods (e.g., debiasing prompts or reweighting) under matched visual conditions.

\subsection{Composition and Statistics}
Each image is associated with structured metadata parsed from prompt text. Table~\ref{tab:prof} shows the profession coverage; Tables~\ref{tab:gender-race}--\ref{tab:age-religion} summarize the labeled social attributes used in the paper.

\begin{table}[t]
\centering
\small
\caption{Profession coverage (single label; $N{=}1{,}200$).}
\label{tab:prof}
\begin{tabular}{lrr}
\toprule
Profession & Count & \% \\
\midrule
Doctor   & 200 & 16.7\% \\
Engineer & 200 & 16.7\% \\
Janitor  & 200 & 16.7\% \\
Lawyer   & 200 & 16.7\% \\
Nurse    & 200 & 16.7\% \\
Teacher  & 200 & 16.7\% \\
\bottomrule
\end{tabular}
\end{table}

\begin{table}[t]
\centering
\small
\caption{Gender and race distributions (single label; $N{=}1{,}200$).}
\label{tab:gender-race}
\begin{tabular}{lrr}
\toprule
\multicolumn{3}{c}{\textbf{Gender}} \\
\midrule
Male         & 498 & 41.5\% \\
Female       & 487 & 40.6\% \\
Unspecified  & 215 & 17.9\% \\
\midrule
\multicolumn{3}{c}{\textbf{Race}} \\
\midrule
Black             & 421 & 35.1\% \\
White             & 403 & 33.6\% \\
South Asian       & 140 & 11.7\% \\
Middle Eastern    & 115 &  9.6\% \\
East Asian        &  86 &  7.2\% \\
Hispanic/Latino   &  19 &  1.6\% \\
Native American   &   7 &  0.6\% \\
Unspecified       &   9 &  0.8\% \\
\bottomrule
\end{tabular}
\end{table}

\begin{table}[t]
\centering
\small
\caption{Age groups and religion (single label; $N{=}1{,}200$).}
\label{tab:age-religion}
\begin{tabular}{lrr}
\toprule
\multicolumn{3}{c}{\textbf{Age group}} \\
\midrule
Young adult  & 343 & 28.6\% \\
Middle-aged  & 374 & 31.2\% \\
Senior       & 357 & 29.8\% \\
Unspecified  & 126 & 10.5\% \\
\midrule
\multicolumn{3}{c}{\textbf{Religion}} \\
\midrule
Christian    & 226 & 18.8\% \\
Muslim       & 230 & 19.2\% \\
Hindu        & 264 & 22.0\% \\
Sikh         &  89 &  7.4\% \\
Buddhist     &  55 &  4.6\% \\
Jewish       &  10 &  0.8\% \\
Taoist       &   6 &  0.5\% \\
Unspecified  & 320 & 26.7\% \\
\bottomrule
\end{tabular}
\end{table}

\paragraph{Exploratory contextual diversity.}
As with \textsc{DIVERSIFY}, some prompts also include broader contextual variation intended to diversify depictions across scenes and settings. We do not treat these prompt-level descriptors as validated image labels, and we do not evaluate them as a separate target dimension.

\paragraph{Label provenance.}
All reported labels are \emph{text-derived} from prompts; no manual annotation of the rendered images was performed. Any broader contextual descriptors that appear in prompt templates are treated, if retained, as prompt metadata rather than image-level ground truth.

\subsection{Data Generation and Quality Control}
\textbf{Prompting scheme.} We start from a profession-specific base prompt (e.g., \emph{``Base prompt (doctor)''}) and create controlled variants (e.g., \emph{V1p1(doctor)}, \emph{V1p2(doctor)}) to toggle a single attribute (gender, race, age group, religion). \textbf{Parsing and validation.} We deterministically extract labels from prompt text using pattern matching over explicit descriptors (\textit{male/female}, \textit{Black/White/East Asian/\dots}, \textit{young adult/middle-aged/senior}, \textit{Christian/Muslim/Hindu/\dots}) and canonicalize them to a closed taxonomy. We perform consistency checks to ensure exactly one profession per prompt and at most one label per single-label attribute.

\subsection{Recommended Uses and Tasks}
This dataset supports: (1) profession-conditioned fairness probes (e.g., are nurses depicted differently by gender or race?); (2) auditing retrieval/captioning models for spurious attribute leakage; and (3) counterfactual testing by swapping a single attribute while holding the rest constant.

\subsection{Ethical Considerations and Risks}
\textbf{Synthetic images are not people.} Demographic labels describe \emph{intended} depictions in generated images; they must not be conflated with real identities. \textbf{Stereotype risk.} Some attribute--profession combinations may mirror societal stereotypes. We encourage reporting performance disaggregated by \emph{both} profession and demographic attribute, and avoiding aggregate-only metrics. \textbf{Context sensitivity.} Broader contextual descriptors used during prompt construction are reductive and should not be interpreted as validated social identity labels.

\subsection{Limitations}
(i) Labels are derived from text prompts rather than human verification of image content; the generator may fail to render the intended attribute. (ii) Taxonomies are coarse and not exhaustive; some identities are absent or underrepresented. (iii) Broader contextual variation reflects prompt design rather than validated image-level labels. (iv) The visual style and biases of the underlying image generator may introduce artifacts unrelated to the prompts.

\subsection{Licensing and Availability}
We release the prompts and metadata (\texttt{.csv/.json}) under a permissive license and provide scripts to regenerate the images where permitted by the image model's license.\footnote{Exact license text and regeneration instructions included in the project repository.} To facilitate reproducibility, we include prompt IDs, attribute labels, and (where applicable) generation seeds.

\subsection{Reproducibility Checklist}
\begin{itemize}\setlength\itemsep{0.2em}
\item \textbf{Files:} \texttt{prompts.txt}, \texttt{metadata.csv} (\texttt{id}, \texttt{profession}, \texttt{gender}, \texttt{race}, \texttt{age\_group}, \texttt{religion}), and a datasheet (\texttt{.xlsx}).
\item \textbf{Counts:} $N{=}1{,}200$ images; 6 professions with 200 images each.
\item \textbf{Determinism:} Prompt $\rightarrow$ label mapping is rule-based and released; image seeds provided when licensing allows.
\item \textbf{Splits:} Deterministic train/dev/test split seeds provided for reproducible evaluation.
\end{itemize}

\section{FairJudge Prompt Templates}
\label{app:prompts}

\begingroup
\setlength{\linewidth}{\columnwidth}
\setlength{\hsize}{\columnwidth}
\sloppy
\setlength{\emergencystretch}{2em}
\setlength{\fboxsep}{6pt}
\setlength{\fboxrule}{0.4pt}

All prompts enforce closed label sets, require visible evidence, and allow a neutral fallback
(\textsc{unspecified}/\textsc{neutral}). We parse only the structured fields during scoring;
rationales are logged for audit.

\vspace{2pt}
\noindent\textbf{Attribute prediction (single prompt shared across models).}

\noindent\captionof{lstlisting}{Attribute prediction prompt (label-constrained with abstention).}
\label{lst:attr-prompt}

\noindent\fbox{%
  \begin{minipage}{\columnwidth}
  \ttfamily\scriptsize
You are given an image of a person. Classify the following using ONLY visible evidence.
If evidence is insufficient for any field, output "unspecified".

Labels:
- Gender: \{male, female, unspecified\}
- Race: \{Black, White, East Asian, South Asian, Middle Eastern,
         Latino\_Hispanic, Indigenous, unspecified\}
- Age: \{child, young adult, middle-aged, elderly, unspecified\}
- Religion: \{Christian, Muslim, Hindu, Buddhist, Jewish, Sikh, Shinto,
             Neutral/unspecified\}
- Disability: \{mobility impairment, blind/low vision, deaf/hard of hearing,
               dwarfism, vitiligo, unspecified\}

Output STRICT JSON (no extra text), with these keys exactly:
\{"Gender": "...", "Race": "...", "Age": "...",
 "Religion": "...", "Disability": "..."\}
  \end{minipage}%
}

\vspace{6pt}
\noindent\textbf{Alignment judging (rubric $\rightarrow$ scalar $[-1,1]$).}

\noindent\captionof{lstlisting}{Prompt--image alignment rubric (later rescaled to $[-1,1]$).}
\label{lst:align-prompt}

\noindent\fbox{%
  \begin{minipage}{\columnwidth}
  \ttfamily\scriptsize
Task: Rate how well the image matches the description.

Description:
"<CAPTION>"

Use this 1--5 rubric:
1 = Not a match at all
2 = Significant discrepancies
3 = Several minor discrepancies
4 = A few minor discrepancies
5 = Matches exactly

Answer in STRICT JSON (no extra text):
\{"Alignment analysis": "<<=100 words, evidence-based>>",
 "Alignment score": "<integer 1-5>"\}
  \end{minipage}%
}

\vspace{6pt}
\noindent\textbf{Profession accuracy + alignment (optional combined prompt).}

\noindent\captionof{lstlisting}{Combined profession + alignment judging.}
\label{lst:prof-align-prompt}

\noindent\fbox{%
  \begin{minipage}{\columnwidth}
  \ttfamily\scriptsize
You are given an image and a candidate profession. First, decide if the image shows that profession.
Then rate alignment using the 1--5 rubric above. Base both decisions ONLY on visible evidence.

Input:
Profession = "<PROFESSION>"
Caption    = "a photo of a <PROFESSION>"

Output STRICT JSON:
\{"Profession prediction": "<yes|no|unspecified>",
 "Prediction evidence": "<<=50 words>>",
 "Alignment analysis": "<<=100 words>>",
 "Alignment score": "<integer 1-5>"\}
  \end{minipage}%
}

\vspace{4pt}
{\footnotesize
\noindent\parbox{\columnwidth}{%
\textit{Implementation notes.}
We keep wording identical across models (GPT-4.1, Gemini 1.5 Pro, Llama-4). At evaluation,
we parse the JSON fields only and linearly rescale the 1--5 alignment rating to
$\overline{s}\in[-1,1]$ for comparison with CLIP cosine; minimal template perturbations are
reported in Sec.~\ref{subsec:results-align}.%
}}

\endgroup

\subsection{Computational Cost}
\label{subsec:cost}

We report wall-clock time per image and marginal API cost per image exactly as recorded in our logs.

\begin{table}[t]
\small
\centering
\caption{Cost/speed per image}
\label{tab:cost-simple}
\begin{tabular}{lcc}
\toprule
Model & Time / image & Cost / image \\
\midrule
Llama-4         & 4.73 s & \$0.0817 \\
CLIP            & 0.21 s     & open-source \\
GPT-4.1         & 4.47 s     & \$0.0025 \\
Gemini 1.5 Pro  & 14.42 s    & \$0.0017 \\
\bottomrule
\end{tabular}
\end{table}

CLIP offers unmatched throughput and negligible marginal cost but provides the least calibrated fairness/faithfulness signal for our purposes, motivating comparison with MLLM judges. Among judges, GPT-4.1 achieves a strong quality--cost trade-off (low per-image cost and moderate latency), while Gemini 1.5 Pro is the cheapest per image but slower in practice. Under our setup, Llama-4 is the most costly because we accessed it via the DeepInfra API; if self-hosted, the \emph{marginal} API cost would be zero but GPU-hours would be substantial. A pragmatic deployment follows a \textbf{cascaded judging} design: route routine cases to CLIP or another cheap metric for broad screening, escalate uncertain or fairness-sensitive cases to a cost-effective judge (e.g., GPT-4.1 or Gemini 1.5 Pro), and invoke a second MLLM opinion only for disagreements or high-stakes audits. This preserves most of the accuracy and alignment gains while controlling runtime and cost. We emphasize that FairJudge is implementation-lightweight---training-free, adapter-free, retrieval-free, requiring only a prompt and parser---but it is \emph{not} computationally lightweight compared with CLIP, and should not be used for very large-scale routine scoring where proxy metrics suffice.

Across attributes, non-demographic categories, and alignment, instruction-following judges offer a practical path to \emph{accountable} evaluation: higher accuracy where cues are visible, better grounding where cues are subtle, and stable rankings under small prompt changes.

\end{document}